\documentclass[letterpaper]{article} 
\usepackage{aaai24}  
\usepackage{times}  
\usepackage{helvet}  
\usepackage{courier}  
\usepackage[hyphens]{url}  
\usepackage{graphicx} 
\usepackage{natbib}  
\usepackage{caption} 
\usepackage{amsmath}
\usepackage{xcolor}
\usepackage{bm}
\usepackage{booktabs}
\def\R{{\rm I\hspace{-0.50ex}R}}

\usepackage{algorithm}
\usepackage{algorithmic}

\usepackage{newfloat}
\usepackage{listings}
\DeclareCaptionStyle{ruled}{labelfont=normalfont,labelsep=colon,strut=off} 
\floatstyle{ruled}
\newfloat{listing}{tb}{lst}{}
\floatname{listing}{Listing}
\title{One Step Learning, One Step Review}
\author{
	Xiaolong Huang\textsuperscript{\rm 1}, Qiankun Li\textsuperscript{\rm 2, 3}\thanks{Corresponding authors (qklee@mail.ustc.edu.cn)}, Xueran Li\textsuperscript{\rm 4, 5}, Xuesong Gao\textsuperscript{\rm 1}
}

\affiliations{
	\textsuperscript{\rm 1}School of Artificial Intelligent, Chongqing University of Technology, Chongqing, China\\
	\textsuperscript{\rm 2}Institute of Intelligent Machines, Hefei Institutes of Physical Science, Chinese Academy of Sciences, Hefei, China\\
	\textsuperscript{\rm 3}Department of Automation, University of Science and Technology of China, Hefei, China\\
	\textsuperscript{\rm 4}Institute of Artificial Intelligence, Hefei Comprehensive National Science Center, Hefei, China\\
	\textsuperscript{\rm 5}Anhui University, Hefei, China\\
	
	\{hirox827, xueran.lxr\}@gmail.com, qklee@mail.ustc.edu.cn, xuesonggxs@foxmail.com
}

\usepackage{bibentry}

\begin{document}
	
	\maketitle
	
	\begin{abstract}
		Visual fine-tuning has garnered significant attention with the rise of pre-trained vision models. The current prevailing method, full fine-tuning, suffers from the issue of knowledge forgetting as it focuses solely on fitting the downstream training set. In this paper, we propose a novel weight rollback-based fine-tuning method called OLOR (One step Learning, One step Review). OLOR combines fine-tuning with optimizers, incorporating a weight rollback term into the weight update term at each step. This ensures consistency in the weight range of upstream and downstream models, effectively mitigating knowledge forgetting and enhancing fine-tuning performance. In addition, a layer-wise penalty is presented to employ penalty decay and the diversified decay rate to adjust the weight rollback levels of layers for adapting varying downstream tasks. Through extensive experiments on various tasks such as image classification, object detection, semantic segmentation, and instance segmentation, we demonstrate the general applicability and state-of-the-art performance of our proposed OLOR. 
		Code is available at https://github.com/rainbow-xiao/OLOR-AAAI-2024.
	\end{abstract}
	
	\section{Introduction}
	
	With the rapid advancement of deep learning technology, numerous large-scale image datasets have been established \cite{laion-2b,imagenet-21k,laion-400m}, resulting in many promising pre-trained visual models \cite{openclip, mae, beit}. These pre-trained models can effectively solve related but distinct visual tasks through transfer learning and fine-tuning techniques \cite{AAAI_transfer_1, AAAI_transfer_2}. The fundamental fine-tuning methods are linear probing and full fine-tuning \cite{linearprob}. In linear probing, the pre-trained model's backbone is frozen, and only the head specific to the downstream task is trained. However, this approach often restricts the performance of the pre-trained backbone. On the other hand, full fine-tuning involves training the entire network directly, but it usually leads to knowledge forgetting \cite{de2021continual}. 
	
	Rehearsal methods \cite{r_16, r_49, replay2020, replay2022}, based on the replay mechanism, involve retraining on a subset of stored upstream samples while learning new tasks. However, this approach is quite inefficient. EWC \cite{EWC} proposes a regularization-based fine-tuning method that uses the Fisher information matrix to determine the importance of weight parameters. This helps adjust the parameters between upstream and downstream tasks, reducing forgetting. 
	L2-SP \cite{L2-SP} uses an L2 penalty to restrict the updates of parameters, addressing knowledge forgetting during fine-tuning. However, it is not compatible with adaptive optimizers \cite{AdamW, AdamW_2}, which may produce the wrong regularization direction.
	Parameter isolation methods \cite{VPT, VPT_cvpr} create new branches or modules for different network models and tasks for downstream tasks. However, it introduces extra new training parameters, requires certain training skills, and has lower generality than rehearsal methods.
	
	In this paper, we propose a novel fine-tuning method combined with optimizers to solve knowledge forgetting, called OLOR (One step Learning, One step Review). Specifically, OLOR introduces a weight rollback term to the weight update term during the fine-tuning stage, allowing the model to gradually approach the pre-trained weights while learning the downstream task. 
	This process avoids delay defects and makes the weights of the upstream and downstream models more similar. 
	In addition, a layer-wise penalty is devised to employ penalty decay and the diversified decay rate to adjust the weight rollback levels of layers. Penalty decay combines feature pyramids with transfer learning, giving more significant weight rollback to shallow layers related to shallow features such as color and texture, and smaller weight backtracking to deep layers related to deep features such as semantic information. The diversified decay rate is adjusted to enhance applicability according to the variations between up and downstream tasks. OLOR with layer-wise penalty enables each layer of the model to update according to its needs, resulting in superior extraction of generalized features.
	Finally, OLOR is incorporated into optimizers, thereby introducing negligible extra computational overhead. It also works well with popular optimizers such as Adam \cite{AdamW, AdamW_2} and SGD \cite{sgd}, meeting specific needs under various conditions.
	
	Our OLOR fine-tuning method achieves state-of-the-art performance on ten popular visual task datasets covering general classification, fine-grained classification, long-tail classification, cross-domain classification, object detection, semantic segmentation, and instance segmentation. Validation experiments and ablation analysis demonstrate the performance of OLOR in solving the problem of knowledge forgetting and the rationality of the parameters.
	
	The main contributions can be summarized as follows.
	\begin{itemize}
		\item 
		We propose a novel fine-tuning method OLOR, which cooperates with optimizers to solve the knowledge forgetting issue, thereby improving fine-tuning performance.
		\item The designed weight rollback avoids delay defects by incorporating the current gradient into the penalty term, thereby correcting the penalty target and smoothing the review process.
		\item A layer-wise penalty is presented to employ penalty decay and the diversified decay rate to adjust the weight rollback levels of layers for adapting varying downstream tasks.
		\item The proposed method achieves state-of-the-art performance on extensive downstream tasks, including different types of image classification, different pre-trained models, and image detection and segmentation. 
	\end{itemize}

	\section{Related Work}
	\subsection{Pre-Training Resource}
	With the rapid advancement of computer vision, numerous large-scale datasets \cite{imagenet-21k,laion-400m, laion-2b} and pre-trained models have emerged. These upstream pre-trained models possess rich features and hold great potential for transferability to other specific downstream tasks.
	ImageNet-21K \cite{imagenet-21k} is the most popular large-scale dataset with over 14 million images, and most networks are pre-trained on it. 
	Recently, a groundbreaking development has taken place with the release of LAION-2B \cite{laion-2b}. This dataset now reigns as the largest, comprising over 2 billion image-text pairs. 
	Then many pre-trained models have been proposed, such as OpenClip \cite{openclip}, BEiT \cite{beit-v2}, MAE \cite{mae}, and EVA \cite{eva-02}. It is worth noting that most of these models' backbones are built upon the foundations of ViT \cite{vit} and ConvNeXt \cite{convnext}.
	
	\subsection{Fine-Tuning Method}
	The process of fine-tuning usually faces an issue known as knowledge forgetting \cite{catastrophic_forgetting_2}. It refers to the model's loss of pre-training learned representations during fine-tuning \cite{mosbach2020stability}. This leads to reduced accuracy on both the upstream and downstream tasks, as the model cannot effectively utilize its potential knowledge \cite{de2021continual,vander2023using}.
	
	To solve this issue, there are currently three categories of approaches, i.e., replay methods, regularization methods, and parameter isolation methods.
	Replay involves periodically training on a subset of upstream task data, thereby retaining knowledge of previous tasks and balancing old and new information \cite{r_16, r_49, replay2020,replay2022}. However, storing and managing updtream task data pose challenges in terms of efficiency, particularly in the contemporary era of massive datasets \cite{laion-2b,li2023lsdir}.
	Regularization-based methods employ techniques such as the fisher information matrix \cite{EWC}, weight decay \cite{kumar2022fine}, and L2 penalty \cite{L2-SP} to restrict parameter updates during fine-tuning. However, these techniques may not be entirely adequate in completely preventing knowledge forgetting. Moreover, the presence of adaptive optimizers \cite{AdamW, AdamW_2} can occasionally impact the direction of regularization \cite{L2-SP}.
	Parameter isolation methods incorporate specific branches or modules into the pre-trained network during downstream fine-tuning, aiming to achieve knowledge transfer through these new modules \cite{VPT, VPT_cvpr, wang2023few}. However, architectural modifications introduce new training parameters and intricate designs. Moreover, training tricks play a crucial role in the effectiveness of the new module, often necessitating multiple rounds of freezing and unfreezing.
	
	To achieve a general and concise fine-tuning method to address knowledge forgetting, the proposed OLOR fine-tuning method combines weight rollback and optimizers to adjust the range of parameter updates. This allows for enhancing pre-trained model representations to improve downstream fine-tuning performance.
	
	\begin{figure*}
		\centering
		\includegraphics[width=0.9\textwidth]{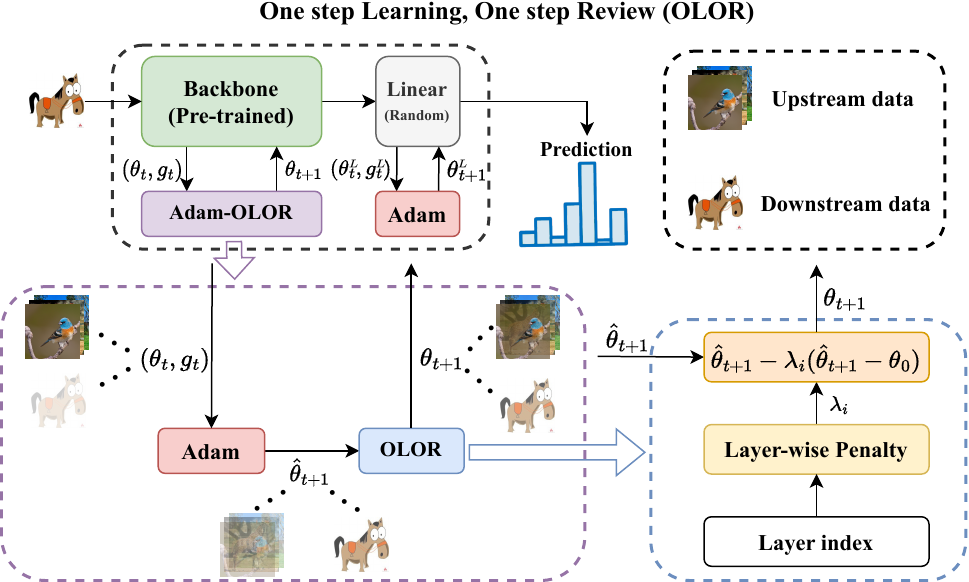}
		\caption{Overview of OLOR using Adam as optimizer, where $\lambda_i$ represents the penalty factor of $i_{th}$ layer, $\theta_t$ and $\hat \theta_{t+1}$ represents the weight and the estimation of next weight (pre-weight) at timestep $t$, respectively. The transparency of the image indicates the knowledge forgetting level.}
        \label{fig:overview}
	\end{figure*}
	
	\section{Method}
	We propose a One step Learning, One step Review (OLOR) method to reduce knowledge forgetting for fine-tuning. OLOR can be seamlessly applied to various downstream tasks among with different optimizers and models. The overall framework is illustrated in Figure \ref{fig:overview}, and detailed pipelines incorporating SGD and Adam are described in Algorithm \ref{algo_sgd} and Algorithm \ref{algo_adam}. This section introduces the delay defect of the previous regularization method, followed by detailed explanations of the OLOR method, which comprises weight rollback and layer-wise penalty.
	
	\definecolor{newcolor}{rgb}{0.8,1,1}
	\newcommand{\weightbackcolor}{Thistle}
	\newcommand{\weightback}[1]{\colorbox{\weightbackcolor}{$\displaystyle #1$}}
	\newcommand{\weightbacktext}[1]{\colorbox{\weightbackcolor}{#1}}
	
	\begin{algorithm}[tb!]
		\caption{OLOR for SGD with Momentum}
		\footnotesize
		\label{algo_sgd}
		\begin{algorithmic}[1]
			\STATE{\textbf{input}: \\ $\eta \in \R$: Initial learning rate, $\beta \in [0, 1)$: momentum factor, $\theta_0$: pre-trained weight, $\iota_1,\iota_2 \in [0, 1], \iota_1 \geq \iota_2$: max and min level of weight rollback respectively, $\gamma \in \R$: weight rollback power}
			\STATE{\textbf{initialize}: \\$t \leftarrow 0$: time step, $m_0 \leftarrow 0$: initial moment vector, $d_0 \leftarrow 0$: initial discrepancy value, $\lambda_i \leftarrow f(\lambda, i, n, \iota_1, \iota_2) / \eta$: calculate penalty factor $\lambda_i$ through $\lambda_i = f(\lambda, i, n, \iota_1, \iota_2) = \iota_2+(1-\frac{i}{n})^\gamma(\iota_1-\iota_2)$, then scale it by dividing $\eta$ to eliminate the scale issue.}
			\REPEAT
			\STATE{$t \leftarrow t + 1$}
			\STATE{$\eta_t \leftarrow$LRScheduler$(\eta_{t-1})$ \hfill(Calculate $\eta_t$ at timestep $t$)}
			\STATE{$g_t \leftarrow \nabla_\theta f_t(\theta_{t-1})$ \hfill(Get batch gradient at timestep $t$)} 
			\STATE{$m_t \leftarrow \beta m_{t-1} + (1 - \beta) g_t $\hfill(Compute momentum)}
			\STATE{$\theta_t \leftarrow \theta_{t-1} - \eta_t \lambda_i d_{t-1} - (1-\eta_t \lambda_i)\eta_t m_t$\hfill(Update weight)}
			\STATE{$d_t \leftarrow (1-\eta_t \lambda_i)(d_{t-1}-\eta_t m_t)$\hfill(Update discrepancy)}
			\UNTIL{Stopping condition is met}
			\RETURN{Parameters $\theta_t$}
		\end{algorithmic}
	\end{algorithm}
	
	\begin{algorithm}[tb!]
		\caption{OLOR for Adam}
		\footnotesize
		\label{algo_adam}
		\begin{algorithmic}[1]
			\STATE{\textbf{input}: \\ $\eta \in \R$: Initial learning rate, $\beta_1, \beta_2 \in [0, 1)$: Exponential decay rates for the moment estimates, $\epsilon$: bias, $\theta_0$: pre-trained weight, $\iota_1,\iota_2 \in [0, 1], \iota_1 \geq \iota_2$: max and min level of weight rollback respectively, $\gamma \in \R$: weight rollback power}
			\STATE{\textbf{initialize}: \\$t \leftarrow 0$: time step, $m_0 \leftarrow 0$: initial first moment vector, $v_0 \leftarrow 0$: initial second moment vector, $d_0 \leftarrow 0$: initial discrepancy value, $\lambda_i \leftarrow f(\lambda, i, n, \iota_1, \iota_2) / \eta$: calculate penalty factor $\lambda_i$ through $\lambda_i = f(\lambda, i, n, \iota_1, \iota_2) = \iota_2+(1-\frac{i}{n})^\gamma(\iota_1-\iota_2)$, then scale it by dividing $\eta$ to eliminate the scale issue.}
			\REPEAT
			\STATE{$t \leftarrow t + 1$}
			\STATE{$\eta_t \leftarrow$LR\_Scheduler$(\eta_{t-1})$ \hfill(Calculate $\eta_t$ at timestep $t$)}
			\STATE{$g_t \leftarrow \nabla f_t(\theta_{t-1})$\hfill(Get batch gradient at timestep $t$)}
			\STATE{$m_t \leftarrow \beta_1 m_{t-1} + (1 - \beta_1) g_t $\hfill(Update first moment vector)}
			\STATE{$v_t \leftarrow \beta_2 v_{t-1} + (1 - \beta_2) g^2_t $\hfill(Update second moment vector)}
			\STATE{$\hat{m}_t \leftarrow m_t/(1 - \beta_1^t)$}
			\STATE{$\hat{{v}}_t \leftarrow v_t/(1 - \beta_2^t)$}
			\STATE{$\theta_t \leftarrow \theta_{t-1} - \eta_t \lambda_i d_{t-1} - \frac{(1-\eta_t \lambda_i)\eta_t \hat{m}_t}{(\sqrt{\hat{v}_t} + \epsilon)}$\hfill(Update weight)}
			\STATE{$d_t \leftarrow (1-\eta_t \lambda_i)(d_{t-1}-\frac{\eta_t \hat{m}_t}{(\sqrt{\hat{v}_t} + \epsilon)})$\hfill(Update discrepancy)}
			\UNTIL{ \textit{stopping criterion is met} }
			\RETURN{optimized parameters $\bm{\theta}_t$}
		\end{algorithmic}
	\end{algorithm}
	
	\subsection{Previous Regularization Mechanisms Have a Delay Defect}
	The implementation of OLOR is inspired by L2 regularization and weight decay, which are popular methods used to regularize the model parameters. However, our findings indicate that their effectiveness does not align with the initial expectation. In the case of the classic SGD optimizer, L2 regularization can be regarded as equivalent to weight decay \cite{AdamW}, which can be defined as follows:
	\begin{equation}
	\theta_t = (1-\lambda)\theta_{t-1} - \eta_t g_t,
	\end{equation}
	where $\theta_t$ represents the model weights at iteration $t$, and $\theta_{t-1}$ is corresponding weights from the previous iteration. $\lambda$ is the regularization factor (weight decay strength). $\eta_t$ is the learning rate at iteration $t$. $g_t$ is the batch gradient computed from the loss function at iteration $t$.
	Weight decay penalizes the weights obtained from the previous iteration by pushing them toward 0. However, in practice, $\lim_{\lambda\to1} \theta_t = - \eta_t g_t$, the weights tend to be pushed towards the negative value of the current gradient instead of 0. This behavior may be different from the initial expectation.
	Furthermore, applying weight decay can actually increase the current weight compared to not applying it. This can be seen in the following inequality:
	\begin{eqnarray}
	(\theta_{t-1} - \eta_t g_t - \lambda \theta_{t-1})^2 > (\theta_{t-1} - \eta_t g_t)^2, 
	\end{eqnarray}
	simplified as: \\
	$$\begin{cases}
	\eta g_t < (1-\frac{\lambda}{2}) \theta_{t-1}, & \text{if } \theta_{t-1} < 0,    \\
	\eta g_t > (1-\frac{\lambda}{2}) \theta_{t-1}, & \text{if } \theta_{t-1} > 0.
	\end{cases}$$
	If $\eta$, $g_t$, $\lambda$, and $\theta_{t-1}$ are in above conditions, using weight decay will drive the current weight away from 0, which is opposite to its target.
	Similarly, this issue with the decay effect also exists in other regularization mechanisms such as L1 regularization, L2-SP, and similar methods.

	\subsection{Weight Rollback}
	The proposed weight rollback is a real-time regularization method that closely follows each weight update step. It aims to bring the current model weights closer to the pre-trained weights to perform knowledge reviewing.
	Specifically, the first step is to calculate the pre-weight $\theta_{pre}$ by gradient:
	\begin{equation}
	\label{eq3}
	\theta_{pre} = \theta_{t-1} - \eta_t g_t ,
	\end{equation}
	where $\theta_{t-1}$ represents the model weights from the previous time step, $\eta_t$ is the learning rate at the current time step, and $g_t$ denotes the gradient.
	Subsequently, the discrepancy $\Delta d$ between $\theta_{pre}$ and the pre-trained weight $\theta_0$ is computed as:
	\begin{equation}
	\label{eq4}
	\Delta d = \theta_{pre} - \theta_0 .
	\end{equation}
	Finally, the weight update process incorporates $\Delta d$, resulting in the adjusted model weights $\theta_t$:
	\begin{equation}
	\label{eq5}
	\theta_t = \theta_{t-1} - \eta_t g_t - \lambda \Delta d .
	\end{equation}
	By substituting Eq. \ref{eq3} and Eq. \ref{eq4} into Eq. \ref{eq5}, we obtain:
	\begin{equation}
	\label{eq6}
	\theta_t = (1 - \lambda)(\theta_{t-1} - \eta_t g_t) + \lambda \theta_0 .
	\end{equation}
	This Eq. \ref{eq6} ensures that $\lim_{\lambda\to1} \theta_t = \theta_0$, which aligns with our expectation and prevents abnormal scenarios. In addition, as the gradient $g_t$ is also subject to a penalty, this process may potentially help to mitigate gradient explosions.
	
	In summary, the weight rollback technique moderates the deviation between $\theta_t$ and $\theta_0$ at each step, thereby alleviating overfitting to the current task and knowledge forgetting to the previous task. 
	
	\subsection{Layer-Wise Penalty}
	\subsubsection{Penalty Decay.}
	For deep learning neural networks, each layer can be conceptualized as a function that processes its input. Given a layer index $i$, this process can be described as follows:
	\begin{equation}
	x_{i+1} = f_i(x^*_i),
	\end{equation}
	where the $f_i$ represents the $i_{th}$ layer. Let $x^u_i$ denotes the input of $f_i$ in upstream tasks with a distribution of $q_i(x^u_i)$, and $x^d_i$ denotes the input of $f_i$ in downstream tasks with a distribution of $p_i(x^d_i)$. Since $q_i(x^u_i)$ are always different from $p_i(x^d_i)$, we first unfreeze all layers to secure $f_i$ will have sufficient update to handle such gap better.
	
	In the study of image feature extraction, a prevailing understanding is that shallow layers are primarily responsible for capturing superficial features \cite{lin2017feature} such as color, texture, and shape. In contrast, deeper layers focus on extracting more profound features like semantic information. This implies that shallow layers are closely linked to the distribution of the data, whereas deep layers are more aligned with task-specific objectives.
	A foundational assumption underlying transfer learning is that $q_i(x^u_i)$ bears a degree of similarity to $p_i(x^d_i)$.
	Consequently, shallow layers tend to exhibit similarities in both pre-training and fine-tuning stages. Additionally, shallow layers require fewer updates compared to their deeper counterparts.
	
	Based on these observations, we propose a layer-wise penalty decay mechanism for weight rollback. This approach gradually reduces the rollback level as the layer depth increases. This strategy encourages shallow layers to extract more general features in downstream tasks while preserving the overall model capacity. For any layer at index $i$, the penalty factor $\lambda_i$ is computed using the following formula:
	
	\begin{equation}
	\lambda_i = \iota_2+(1-\frac{i}{n})(\iota_1-\iota_2),
	\end{equation}
	where $n$ represents the total number of layers in the pre-trained model, $\iota_1$ and $\iota_2$ denote the maximum and minimum rollback levels, respectively.
	
	\subsubsection{Diversified Decay Rate.}
	
	Across various downstream tasks, the target objectives often exhibit varying degrees of dissimilarity from those of the upstream task. To accommodate this variability, we propose adjusting the rate of penalty decay between layers by introducing a power exponent $\gamma$ to the weight rollback value. Mathematically, this adjustment can be expressed as:
	\begin{equation}
	1-\frac{i}{n} \longrightarrow{} (1-\frac{i}{n})^\gamma.
	\end{equation}
	This dynamic adjustment helps mitigate the bias stemming from a fixed rate decay of the similarities between $q_i(x^u_i)$ and $p_i(x^d_i)$ across different layer indices $i$.
	Consequently, the penalty decay becomes more adaptable and versatile, catering to a spectrum of requirements dictated by the various downstream tasks.
	
	\begin{table}
		\centering
		\begin{tabular}{llll}
			\hline
			Dataset         & Images & Categories & Type \\
			\hline
			CIFAR-100   &   60,000 & 100     & General \\
			SVHN         & 600,000 & 10     & General \\
			CUB-200   & 11,788  & 200  & Fine-grained \\
			Stanford Cars  & 16,185 & 196 &  Fine-grained \\
			Places-LT    & 62,500 & 365  &  Long-tailed \\
			IP102           & 75,222  & 102  &  Long-tailed \\
			OfficeHome      & 15,500   & 4 $\times$ 65 &  Cross-domain \\
			PACS           & 9,991  & 4 $\times$ 7 &  Cross-domain \\
			COCO2017           & 163,957  & 80 &  Detection \\
			ADE20K           & 27,574  & 3688 &  Segmentation \\
			\hline
		\end{tabular}
        \caption{Details of the fine-tuning datasets.}
		\label{tab:dataset}
	\end{table}
	
	\begin{table*}
		\centering
		\small
		\begin{tabular}{lcccccccc}
			\toprule
			& \multicolumn{2}{c}{General (ID)} & \multicolumn{2}{c}{Fine-Grained (ID)} & \multicolumn{2}{c}{Long-Tailed (OOD)} & \multicolumn{2}{c}{Cross-Domain (OOD)} \\
			\cmidrule(lr){2-3} \cmidrule(lr){4-5} \cmidrule(lr){6-7} \cmidrule(lr){8-9}
			Method & Cifar-100 & SVHN & CUB-200 & StanfordCars & Places-LT & IP102 & OfficeHome & PACS \\
			\midrule
			\multicolumn{3}{l}{ViT-B Backbone}\\
			\cmidrule{1-1}
			Linear & 72.50 & 58.79 & 75.01 & 38.03 & 31.95 & 64.93 & 79.96 & 71.88\\
			Full & 87.76 & 97.27 & 81.34 & 75.55 & 31.59 & 74.09 & 84.39 &	87.79\\
			L2-SP & 88.17 &	97.12 &	81.65 &	75.55 &	31.22 &	73.75 &	84.74 &	87.74 \\
			VPT & 91.49 & 94.37 & 81.86	& 58.24 & 37.02	 & 70.41 &	86.48 &	77.44\\
			\textbf{OLOR-Adam (ours)} &	\textbf{92.89} & \textbf{97.35} & \textbf{84.84} & \textbf{82.02} &	\textbf{38.07} &	\textbf{75.34} & \textbf{89.05} & \textbf{94.38}\\
			\midrule
			\multicolumn{3}{l}{ConvNeXt-B Backbone}\\
			\cmidrule{1-1}
			Linear & 81.70&	69.21&	87.85&	50.21&	36.41&	70.77&	92.40&	93.46\\
			Full & 92.72&	96.97&	88.59&	88.67&	38.61&	75.01&	91.78&	95.51\\
			L2-SP & 92.84&	97.01&	88.82&	88.83&	38.52&	75.20&	90.61&	95.90 \\
			VPT & 88.71&	81.58&	87.88&	51.58&	36.32&	71.22&	92.31&	93.75\\
			\textbf{OLOR-SGD (ours)} &\textbf{92.86}&	\textbf{97.12}&	\textbf{89.47}&	\textbf{88.99}&	\textbf{39.36}&	\textbf{75.44}&	\textbf{92.59}&	\textbf{96.63}\\
			\bottomrule
		\end{tabular}
        \caption{Comparison of fine-tuning results on various types of classification datasets (general, fine-grained, long-tailed, cross-domain).}
		\label{tab:class-finetune}
	\end{table*}
	
	\section{Experiments}
	\subsection{Experiment Configuration}
	\subsubsection{Pre-Trained Backbones.}
	The experiments employ CNN-based ConvNeXt \cite{convnext} and Transformer-based Vision Transformers (ViT) \cite{vit} as backbones. For both types of models, pre-trained weights from ImageNet-1K (MAE) \cite{imagenet-1k}, ImageNet-21K (supervised) \cite{imagenet-21k} and LAION-2B (CLIP) \cite{laion-2b} datasets are utilized, where the weights from ImageNet-21K undergoes supervised pre-training, and the others are based on self-supervised pre-training diagram.
	

	\subsubsection{Downstream Tasks.}
	We experiment on ten popular visual task datasets, i.e., CIFAR-100 \cite{cifar-100}, SVHN \cite{svhn}, CUB-200 \cite{cub-200}, Stanford Cars \cite{standfordcars}, Places-LT \cite{places-lt}, IP102 \cite{ip102}, OfficeHome \cite{officehome}, and PACS \cite{PACS}, covering general classification, fine-grained classification, long-tailed classification, cross-domain classification, object detection, semantic segmentation, and instance segmentation. More details are listed in Table \ref{tab:dataset}.

	\subsubsection{Baselines.}
	
	To ensure a comprehensive comparison, we select the state-of-the-art and classic methods as our baselines. 
	These encompass Full Fine-tuning (Full), Linear Probing (Linear) \cite{linearprob}, L2-SP \cite{L2-SP}, and VPT \cite{VPT}.
	Following prior works \cite{cnnsgdvitadam}, CNN-based Backbones are usually combined with the SGD optimizer, while Transformer-based Backbones are paired with the Adam optimizer.

	\subsubsection{Implementation Details.}
	The input image size is set at $224 \times 224$. 
	The batch size varies depending on the freezing strategy. Specifically, 128, 256 and 512 are chosen for full unfreezing, parameter isolated, and full freezing based methods, respectively. Regarding the learning rate, for ConvNeXt backbones, we employ the SGD optimizer with a momentum of 0.9. The learning rates differ based on the freezing strategy. In detail, 1e-2, 2e-2 and 4e-2 for full unfreezing, parameter isolated, and full freezing based methods, respectively. For ViT backbones, we use the Adam optimizer with a momentum of (0.9, 0.999). The learning rates for ViT backbones also vary according to the freezing strategy, i.e., 1e-4 for full unfreezing, 2e-4 for partial unfreezing, and 4e-4 for full freezing. We train on cross-domain datasets for 30 epochs, while for other datasets, we train for 50 epochs.
	The experiments are performed on two A5000 GPU with 24 GB memory and Ubuntu 20.04 operating system. 
	Python 3.8.3 serves as the programming language, while PyTorch 2.0.0 framework is employed. 
	In addition, the source code is openly available on GitHub.

	\subsection{Main Results}
	\label{sec:mainresults}
	\subsubsection{Results on Classification Tasks.}
	To verify the wide adaptability of OLOR on various types of datasets, we conduct a comprehensive comparison with other state-of-the-art fine-tuning methods. We evaluate these methods on 10 popular classification datasets, each showcasing a range of data distributions and characteristics. In addition, the Backbone in the experiment covers ViT-B and ConvNeXt-B, corresponding to Adam and SGD optimizers, respectively. 
	
	The experiment results are listed in Table \ref{tab:class-finetune}. It can be observed that our OLOR achieves a new state-of-the-art on all datasets. Notably, in in-distribution (ID) datasets, OLOR-Adam surpasses the previously leading L2-SP method by an impressive margin of 6.47\% in accuracy. Moreover, when confronted with two more challenging out-of-distribution (OOD) datasets, OLOR-Adam achieves accuracy improvements of 2.57\% and 7.38\%, respectively, outperforming the optimal methods.
	
	Since the pre-trained ConvNeXt model is more stable than the ViT structure, there is not much difference between different methods in fine-tuning. However, our OLOR-SGD still consistently improves fine-tuning accuracy across all datasets. These results demonstrate the robustness and effectiveness of the proposed OLOR in various tasks.

	\subsubsection{Results on Detection and Segmentation Tasks.}
	Due to the complexity of detection and segmentation tasks, most existing fine-tuning methods struggle with applicability and validation. However, integrated with the optimizer, our OLOR approach can easily be applied to these tasks. Table \ref{tab:coco2017} shows the results of object detection and instance segmentation on the COCO2017 dataset, while Table \ref{tab:ade20k} showcases the performance of semantic segmentation on the ADE20K dataset. OLOR consistently outperforms the Baseline by approximately 1\% in all metrics, demonstrating its versatility and effectiveness in more complex detection and segmentation tasks.

	\begin{table}
		\centering
		\small
		\begin{tabular}{lcccc}
			\toprule
			Method & Model & Dataset& $Bbox_m$ & $Segm_m$ \\
			\midrule
			Full & Mask R-CNN & COCO2017 & 40.20 & 36.00 \\
			\textbf{OLOR} & Mask R-CNN & COCO2017 &\textbf{41.10} & \textbf{36.90} \\
			\bottomrule
		\end{tabular}
        \caption{Results of object detection and instance segmentation using the ConvNeXt-B as backbone.}
		\label{tab:coco2017}
	\end{table}

	\begin{table}
		\centering
		\small
		\renewcommand{\tabcolsep}{14pt}
		\begin{tabular}{lccc}
			\toprule
			Method & Model & Dataset& $IOU_m$ \\
			\midrule
			Full & UperNet & ADE20K  & 43.65 \\
			\textbf{OLOR} & UperNet & ADE20K  &\textbf{44.62} \\
			\bottomrule
		\end{tabular}
        \caption{Results of semantic segmentation using the ViT-B as backbone.}
		\label{tab:ade20k}
	\end{table}
	
	\begin{table}
		\centering
		\small
		\renewcommand{\tabcolsep}{9pt}
		\begin{tabular}{lccc}
			\toprule
			Method & Supervised & OpenCLIP & MAE \\
			\midrule
			Linear & 71.88 &	95.61 &	36.72 \\
			Full   & 87.79 &	47.17 &	84.18 \\
			L2-SP &	87.74 &	45.56 &	85.79 \\
			VPT  & 77.44 &	97.46 &	50.54 \\
			\textbf{OLOR (ours)} &	\textbf{94.38} & \textbf{98.10} & \textbf{89.26} \\
			\bottomrule
		\end{tabular}
        \caption{Results of using different pre-trained models on the PACS dataset.}
		\label{tab:vitopenclipmae}
	\end{table}
	
	\subsubsection{Results of Using Different Pre-Trained Models.}
	Considering that the performances of different fine-tuning methods may vary when using different pre-trained models, we further conduct experiments to explore and compare. The pre-trained ViT-B model weights are obtained from ImageNet-21K (supervised), LAION-2B (CLIP), and ImageNet-1K (MAE). The fine-tuning experiments are based on the challenging PACS dataset.
	
	As listed in Table \ref{tab:vitopenclipmae}, our OLOR consistently achieves state-of-the-art results across all pre-trained models. Specifically, OLOR surpasses other leading methods by 5.08\%, 0.64\%, and 3.47\% when using Supervised, CLIP, and MAE, respectively. While other methods struggle to adapt to all pre-trained models simultaneously, our OLOR demonstrates potential across all pre-trained models.

	\subsubsection{Summary of Main Results.}
	In summary, the above experiments show that OLOR achieves SOTA when applied to multiple downstream tasks, utilizing diverse pre-trained backbones. These results demonstrate the generalizability and effectiveness of the OLOR fine-tuning method.

	\subsection{Analysis and Discussion}
	
	\begin{figure}
		\begin{center}
			\centering
			\includegraphics[width=\linewidth]{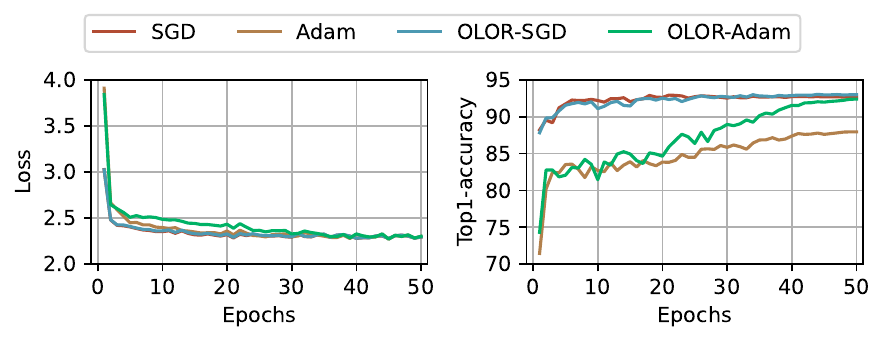}
		\end{center}
		\vspace{-0.4cm}
		\caption{Train loss and valid top1 accuracy on CIfar-100, using ViT-B with Adam and ConvNext-B with SGD.}
		\vspace{-0.2cm}
		\label{fig-loss}
	\end{figure}
	\subsubsection{Compatibility Analysis.}
	As shown in Figure \ref{fig-loss}, adopting weight rollback in different types of models and optimizers generally improves the performance. Due to the restriction on parameters, OLOR leads to slower loss converging speed at first, but ultimately becomes competitive with the full method. According to the validation results, OLOR potentially helps reduce knowledge forgetting, resulting in far superior top1 accuracy, especially when cooperating with Adam applied in Vision Transformers.

	\begin{figure}
		\begin{center}
			\centering
			\includegraphics[width=\linewidth]{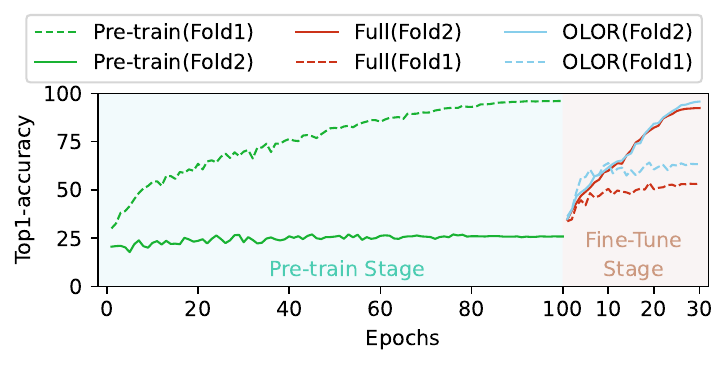}
			\includegraphics[width=\linewidth]{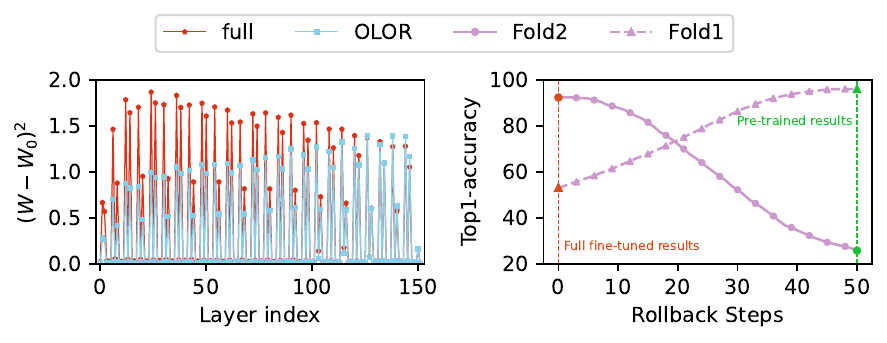}
		\end{center}
		\vspace{-0.4cm}
		\caption{Knowledge forgetting test on PACS. Fold 1 as train set and fold 2 as valid set during pre-training, splits during fine-tuning is opposite to pre-training.}
		\vspace{-0.2cm}
		\label{fig-review}
	\end{figure}
	\subsubsection{Knowledge Forgetting Test.}
	To assess potential knowledge forgetting, we conduct a study on the PACS dataset using ViT-B and Adam. Firstly, split the dataset into two folds, the first fold contains data from three domains, cartoon, photo and sketch respectively, denote as $\mathcal D_1$, the second fold contains data from art painting domain, denote as $\mathcal D_2$. For training stage, we first pre-train a model using $\mathcal D_1$ as train set and $\mathcal D_2$ as valid set for 100 epochs, then fine-tune the model using $\mathcal D_2$ as train set and $\mathcal D_1$ as valid set for 30 epochs through Full and OLOR methods, the discrepancy between fine-tuned weight $\theta$ and pre-trained weight $\theta_0$ using different methods are recorded. Additionally, we perform zero-shot reviewing, rolling back full fine-tuned weights to pre-trained weights in 50 steps. Figure \ref{fig-review} reports the results, weight discrepancy is generally much smaller using OLOR, when setting max rollback level $\iota_1$ to 0.01, rollback power $\gamma$ to 1, OLOR not only performs well in knowledge reviewing, but also benefits for current learning. And the zero-shot reviewing result shows weight rollback itself is indeed a helpful method for just reviewing.
	
	\begin{figure}
		\begin{center}
			\centering
			\includegraphics[width=\linewidth]{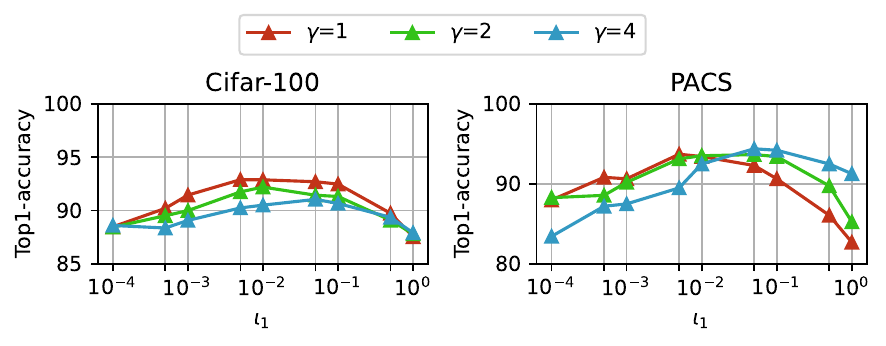}
		\end{center}
		\vspace{-0.4cm}
		\caption{Hyper-parameters exploring experiments on Cifar-100(left) and PACS(right), both using ViT-B with Adam.}
		\vspace{-0.2cm}
		\label{fig-hyp}
	\end{figure}
	\subsubsection{Hyper-Parameter Exploration.}
	We conduct experiments on Cifar-100(ID) and PACS(OOD) to study the appropriate hyper-parameters for different types of tasks. Deep layers usually require significant updates to effectively extract features related to the downstream task, thus we set the min rollback level $\iota_2$ to 0 by default to simplify hyper-parameter settings, for max rollback level $\iota_1$, we search from \{0.0001, 0.0005, 0.001, 0.005, 0.01, 0.05, 0.1, 0.5, 1\}, for weight rollback power $\gamma$, we search from \{1, 2, 4\}. Figure \ref{fig-hyp} shows the findings. We suggest applying small power if the task target of the fine-tuning stage is similar to the pre-training stage, and large max rollback level if the data distribution of downstream task is similar to upstream task.

	\begin{figure}
		\begin{center}
			\includegraphics[width=0.48\textwidth]{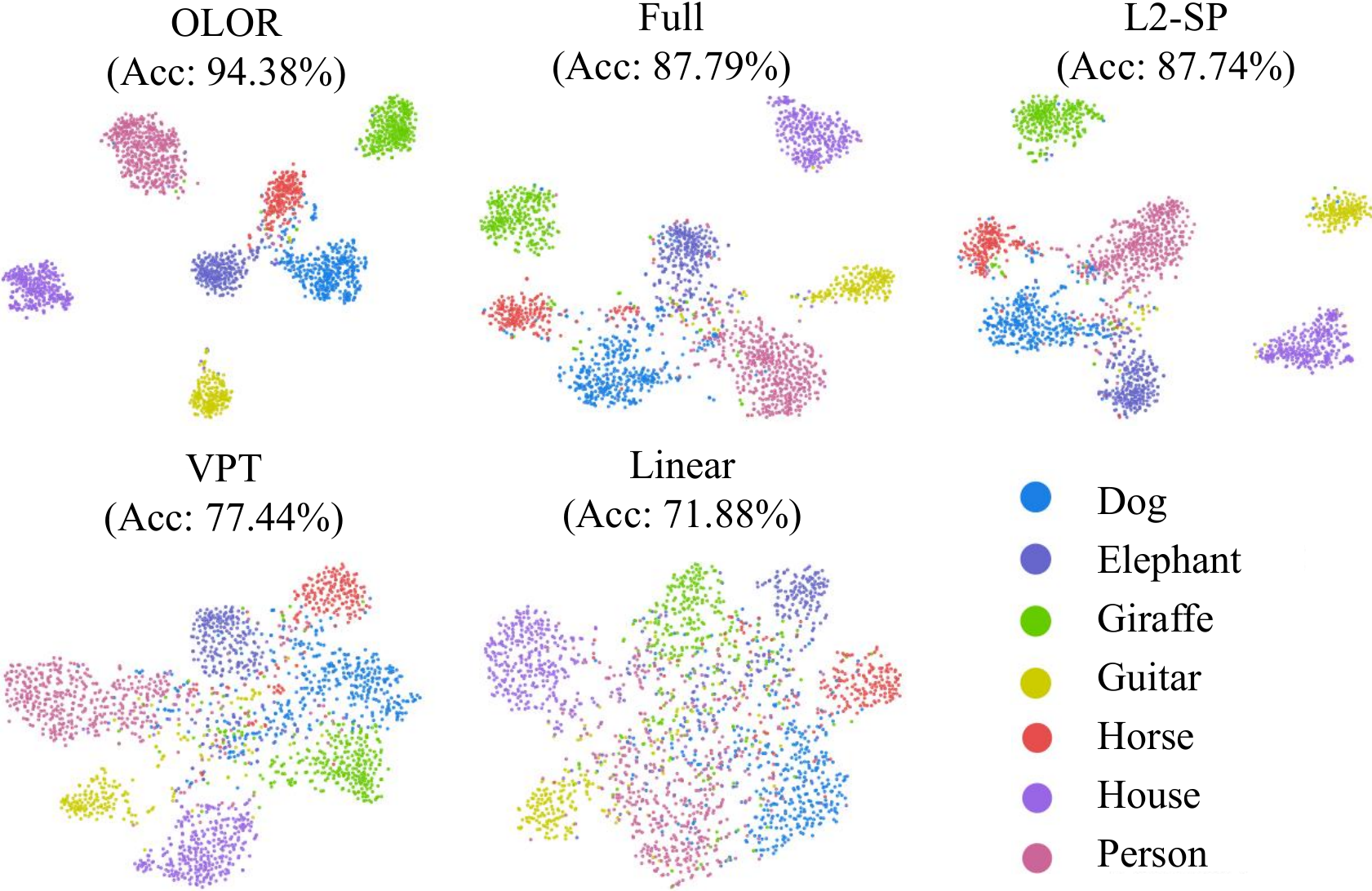}
		\end{center}
		\vspace{-0.4cm}
		\caption{Feature visualization on PACS test set. We use features extracted by backbone to perform t-SNE visualization, and the Top1-accuracy are reported additionally.}
		\vspace{-0.2cm}
		\label{fig-visual}
	\end{figure}
	\subsubsection{Feature Visualization.}
	We visualized the feature distributions for all methods on PACS test set through t-SNE to evaluate the quality of the extracted features. Experiments are based on ViT-B and Adam. As shown in Figure \ref{fig-visual}, compared with previous methods, OLOR generally separates the representation vectors of different classes much better, demonstrating superior ability on representation.
	
	\section{Conclusions}
	In this paper, we propose a novel fine-tuning method named OLOR to solve the challenge of knowledge forgetting in neural networks. OLOR encompasses weight rollback and layer-wise penalty. OLOR incorporates the weight rollback term into the weight update term at each step, and can be implemented in popular optimizers. This operation allows the model to gradually approach the pre-trained weights while learning the downstream task, making the weights of the upstream and downstream models more similar.  In addition, the layer-wise penalty employs penalty decay and the diversified decay rate to adjust the weight rollback levels of layers for adapting varying downstream tasks. Our OLOR achieves state-of-the-art performance on extensive downstream tasks. Validation experiments and ablation analysis demonstrate the effectiveness of the proposed method.
	
	\section{Additional Implementation Details}
	In the Main Results section, when conducting experiments on various downstream tasks, OLOR utilizes the hyper-parameter configurations listed in Table \ref{tab:configresults}. 
	For experiments involving different pre-trained models, the hyper-parameter configurations for OLOR are listed in Table \ref{tab:configresu}.
	
	\begin{table}
		\centering
		\begin{tabular}{lcccccc}
			\toprule
			Datasets & \multicolumn{3}{c}{ViT-Based} & \multicolumn{3}{c}{CNN-based} \\
			\cmidrule(lr){2-4} \cmidrule(lr){5-7}
			& $\iota_1$ &$\iota_2$ & $\gamma$ & $\iota_1$ &$\iota_2$ & $\gamma$ \\
			\midrule
			Cifar-100 & 5e-3 & 0 & 2 & 5e-3 & 0 &2\\
			SVHN   & 5e-3 &0 & 2  & 1e-4 &0 &2\\
			CUB-200 &5e-2 &	0 &	2& 1e-2 &0 &2\\
			StanfordCars & 1e-2 & 0& 4 & 1e-4 & 0 & 2\\
			Places-LT & 1e-1 &0 &4 & 1e-2 & 0 &4\\
			IP102  & 1e-1&0&1 &5e-3 & 0 &1\\
			OfficeHome &1e-2 &0&1 & 1 &0 &1\\
			PACS & 1e-1&0&4 & 5e-2 &0 &4\\
			COCO2017 & -&-&- & 1e-2 & 0 &2\\
			ADE20K& 1e-4&0&1 & - & - &-\\
			\bottomrule
		\end{tabular}
        \caption{Hyper-parameter configuration of OLOR for different downstream tasks.}
		\label{tab:configresults}
	\end{table}
	
	\begin{table}
		\centering
		\begin{tabular}{lcccccc}
			\toprule
			Pre-trained Method  & $\iota_1$ &$\iota_2$ & $\gamma$ \\
			\midrule
			Supervised  & 1e-2 &0 & 2 \\
			OpenCLIP &1e-2 &	0 &	2\\
			MAE & 1e-2 & 0& 2\\
			\bottomrule
		\end{tabular}
        \caption{Hyper-parameter configuration of OLOR for different pre-trained models.}
		\label{tab:configresu}
	\end{table}

	\section{Acknowledgments}
	This work was supported by the Students' Innovation and Entrepreneurship Foundation of USTC (No.XY2023S007).
	We would like to sincerely appreciate the anonymous reviewers for their valuable suggestions that helped us to improve this paper.

\bibliography{aaai24}

\end{document}